\pdfoutput=1
\documentclass[11pt]{article}

\usepackage[final]{EACL2023}

\usepackage{times}
\usepackage{comment}
\usepackage{latexsym}
\usepackage{array}
\usepackage{multirow}
\usepackage{color}
\usepackage{pgfplots}
\usepackage{dingbat}
\usepackage{pifont}
\usepackage{contour}
\usepackage{cancel}
\usepackage{subcaption}
\usepackage[normalem]{ulem}

\contourlength{0.8pt}

\usepackage[T1]{fontenc}
\usepackage[utf8]{inputenc}

\usepackage{microtype}

\usepackage{inconsolata}

\title{Discriminative Models Can Still Outperform Generative Models in Aspect Based Sentiment Analysis}

\author{Dhruv Mullick \\
  Dept. of Computing Science \\
  University of Alberta \\
  \texttt{mullick@ualberta.ca} \\\And
  Alona Fyshe \\
  Dept. of Computing Science \\
  University of Alberta \\
  \texttt{alona@ualberta.ca} \\\And
  Bilal Ghanem \\
  Dept. of Computing Science \\
  University of Alberta \\
  \texttt{bilalhgm@gmail.com} \\\\
  }

\begin{document}
\maketitle
\begin{abstract}

Aspect-based Sentiment Analysis (ABSA) helps to explain customers' opinions towards products. Earlier, ABSA models were discriminative, but recently generative models have been used to generate aspects and polarities directly from reviews. Previous results showed that generative models outperform discriminative models on several English ABSA datasets. Here, we evaluate and contrast two state-of-the-art discriminative and generative models in several settings: cross-lingual, cross-domain, and cross-lingual and domain, to understand generalizability in settings other than English mono-lingual in-domain. Our evaluation shows that discriminative models can still outperform generative models in a few settings. Further, we present a problem faced by generative models in cross-lingual settings, and demonstrate a mitigation strategy that combines the best of discriminative and generative models.

\end{abstract}

\section{Introduction}

Online reviews make it easy for customers to share their feelings about products in a quick and efficient way. Companies with millions of customers receive massive amounts of such online reviews that can't be analyzed manually, thus needing automation.

A few languages receive more research effort compared to others (e.g. English vs. Swahili). Although the community has remarkably accelerated the improvement of English NLP techniques, techniques for other languages lag behind. Working on a lower resource language is a challenging task, where few datasets, lexicons, and models exist. Thus, utilizing cross-lingual approaches is important to migrate model's ability across languages.

ABSA involves predicting aspect terms and their associated sentiment polarities \cite{liu_sentiment_2012} - for example, "Service was good at the restaurant, but food was not" has two aspect terms ("service" and "food"), associated with sentiments "positive" and "negative", respectively. Training such an ABSA model requires a suitable amount of data. Hence, in low resource settings, it can be difficult to use ABSA to analyze reviews. A solution is to use models which were trained in some other setting \cite{liu-etal-2020-zero}. For example, to perform ABSA in Swahili, use a model trained on ABSA in English. For such settings, we conduct a comparative study of discriminative and generative ABSA models. 

Discriminative models, which use decision boundaries to make predictions, commonly use sequence labeling techniques to detect aspects in a given review (extraction) and then use another step to classify those aspects (classification). Notably, discriminative models can do both at once, but this does not always improve performance \cite{hu-etal-2019-open}. Instead, generative models use encoder-decoder language models to learn probability distributions over words, and generate aspects and sentiment polarities together without separate extraction and classification steps.

Prior work shows that generative models achieve better performance than discriminative models in the English in-domain setting \cite{zhang-etal-2021-towards-generative, yan-etal-2021-unified}. However, none have explored performance in cross-lingual or cross-domain settings. These settings are of importance to cases involving low resources for a given domain or language.

In our work: \textbf{a)} We evaluate performance of the two model types in various settings by comparing state-of-the-art representatives. Our results demonstrate that discriminative models can still perform better than generative models in a few cases; \textbf{b)} We find that generative models face problems with aspect extraction in cross-lingual settings. As a mitigation, we propose a masking approach which gives large improvements (up to 20\%), by combining a generative model's semantic understanding of labels, along with a discriminative model's constraint of only generating words from the input sentence.

\section{Data}

In our experiments, we considered several languages and domains for a thorough evaluation. For languages we used SemEval datasets - Restaurant (Rest16) \cite{pontiki-etal-2016-semeval} in English, Spanish and Russian. For domains we used Rest16 and Laptop (Lap14) from SemEval \cite{pontiki-etal-2014-semeval}. Additionally, we used the MAMS dataset ~\cite{jiang-etal-2019-challenge} of the restaurant domain. In MAMS, each sentence contains at least two aspects with different polarities, making the dataset more challenging than the SemEval datasets. Further dataset details can be found in Appendix \ref{app:dataset_details}.

\section{Models} \label{cross:models}

We perform experiments on two generative and one discriminative ABSA model. We contrast the generative and discriminative ABSA paradigms by considering a representative model for each, which shows state-of-the-art performance. We also consider a generative model constrained to generate aspects from the input sentence.

\textbf{1) SPAN-MBERT Discriminative model}: we considered SPAN-BERT \cite{hu-etal-2019-open}, which is a state-of-the-art ABSA model that uses a BERT transformer. It has good performance on English mono-lingual in-domain datasets, and has been used as a baseline for generative models \cite{yan-etal-2021-unified}. The model extracts continuous spans of text for multiple target aspect terms and classifies their polarities using contextualised span representations. To use this model in cross-lingual settings, we replace its encoder with a multilingual BERT model. We call this methodology - "SPAN-MBERT".

\textbf{2) mBART Non-Masked Generative model}: we used an encoder-decoder BART-based approach \cite{yan-etal-2021-unified}. This model takes a review as input and generates aspects and their polarities. The aspect-polarity terms have the format: "service positive $<$sep$>$ food negative", indicating presence of two aspect terms ("service" and "food"), with associated polarities ("positive" and "negative"). A separator token "$<$sep$>$" is used to demarcate a separation between the multiple aspect-polarity pairs in a review. To use this approach in cross-lingual settings, we use a multilingual BART model. We call this methodology - "mBART Non-Masked" ("Non-Masked" term explained in Sec. \ref{masked_logits}).

\textbf{3) mBART Masked Generative model}: this model is a modification of the mBART Non-Masked model. It uses masking to constrain the model to generate aspects from the input sentence itself (details in Section \ref{masked_logits}).

The SPAN-MBERT model and both mBART models use a similar mBERT-like encoder, implying that performance differences between them is only due to their discriminative and generative components and not because of the sentence encoding.

\section{Logit Masking \& mBART Masked} \label{masked_logits}

ABSA involves finding aspect terms which are phrases within the input sentence. Generative models, however, are not restricted to generating words from the input sentence. They can generate words from the entire vocabulary, and are hence at a risk of generating out-of-sentence words ("hallucinating") \cite{ji2022survey}. Discriminative models, do not face this problem as they simply label each word in the input sentence as an aspect or non-aspect \cite{li2019unified, li-etal-2019-exploiting}.

This hallucination problem for the generative mBART Non-Masked model is confirmed in the error analysis (Sec. \ref{cross:ea}). A solution to this problem is to constrain the generative model's decoding step so that the model can only generate from a fixed set of tokens that exist in the input sentence \cite{huang-etal-2022-multilingual-generative, de-cao-etal-2022-multilingual}. 

We implemented this solution for mBART by "masking" tokens not in the input sentence. Masking changes a token's softmax inputs to -Infinity before calculating softmax probabilities, ensuring that the generative model can only select a token that appears in the input sentence. We call this the "mBART Masked" model. This model combines the best of other models - a generative model's semantic understanding of labels \cite{zhang-etal-2021-towards-generative}, and a discriminative model's constraint of only generating words from the input sentence.

\section{Experiments and Results}

For the three models mentioned in Section \ref{cross:models}, we ran experiments under different settings. We compare \textbf{1)} the mBART Non-Masked model against the SPAN-MBERT model to check if generative models indeed outperform discriminative models in various settings; \textbf{2)} the mBART Non-Masked model against the mBART Masked model to see if constrained decoding is beneficial.

We ran all experiments with 10 random seeds for robustness. The results are checked for statistical significance using the Yuen-Welch test. We find that the standard deviation in results is high because of degenerate runs. Transformer-based models are known to produce degenerate runs when fine-tuned on small datasets \cite{zhang2020revisiting}.

\subsection{Monolingual and In-Domain}

We evaluated models with train and test data from the same domain and language. As shown in Table \ref{tab:mono-lingual}, the mBART Non-Masked model performs better than the SPAN-MBERT model in all cases except the MAMS case. The mBART Non-Masked model and the mBART Masked models have no statistically significant difference in performance.

\begin{table*}[!htb]
\centering
\begin{subtable}[!htb]{0.8\textwidth}
\centering
\small
\begin{tabular}{l|c|c|c}
\hline
\multirow{1}{*}{\textbf{Domain$_{Lang}$}} & \multicolumn{1}{c|}{\textbf{SPAN-MBERT}} & \multicolumn{1}{c|}{\textbf{mBART Non-Masked}} & \multicolumn{1}{c}{\textbf{mBART Masked}} \\
\hline
\textbf{Rest16$_{En}$}      & 60.96 {\footnotesize ($\pm$ 2.15)}* & \textbf{74.17} {\footnotesize ($\pm$ 2.13)} & \underline{74.02} {\footnotesize ($\pm$ 2.17)} \\
\textbf{Rest16$_{Es}$}      & 64.72 {\footnotesize ($\pm$ 1.19)}* & \textbf{69.83} {\footnotesize ($\pm$ 1.28)} & \underline{69.50} {\footnotesize ($\pm$ 1.34)} \\
\textbf{Lap14$_{En}$}   & 57.20 {\footnotesize ($\pm$ 1.51)}* &  \underline{66.35} {\footnotesize ($\pm$ 2.70)}  & \textbf{66.65} {\footnotesize ($\pm$ 2.21)} \\
\textbf{MAMS$_{En}$}        & \textbf{66.00} {\footnotesize ($\pm$ 0.42)}* & \underline{61.14}  {\footnotesize ($\pm$ 1.20)} & 60.97 {\footnotesize ($\pm$ 1.18)} \\
\textbf{Rest16$_{Ru}$}      & 54.60 {\footnotesize ($\pm$ 2.27)}* & \textbf{68.55} {\footnotesize ($\pm$ 1.28)} & \underline{68.38} {\footnotesize ($\pm$ 1.13)} \\
\hline
\end{tabular}
\caption{Mono-lingual and in-domain F1 scores.}
\label{tab:mono-lingual}
\end{subtable}

\vspace{0.3cm}

\centering
\begin{subtable}[!htb]{0.8\textwidth}
\centering
\small
\begin{tabular}{l|c|c|c}
\hline
\multirow{1}{*}{\textbf{Train $\rightarrow$ Test}} & \multicolumn{1}{c|}{\textbf{SPAN-MBERT}} & \multicolumn{1}{c|}{\textbf{mBART Non-Masked}} & \multicolumn{1}{c}{\textbf{mBART Masked}} \\
\hline
\textbf{$Es$ $\rightarrow$ $En$} & 48.87 {\footnotesize ($\pm$ 2.22)}* & \underline{55.88} {\footnotesize ($\pm$ 14.96)} & 	\textbf{61.90} {\footnotesize ($\pm$ 12.72)}  \\
\textbf{$Ru$ $\rightarrow$ $En$} &	32.89 {\footnotesize ($\pm$ 6.16)} * & \underline{64.77} {\footnotesize ($\pm$ 3.62)} &	\textbf{67.75} {\footnotesize ($\pm$ 3.91)} \\
\hline
\textbf{$En$ $\rightarrow$ $Ru$} &	39.86 {\footnotesize ($\pm$ 1.89)} * & \underline{50.66} {\footnotesize ($\pm$ 8.12)} &	\textbf{54.02} {\footnotesize ($\pm$ 10.84)} \\
\textbf{$Es$ $\rightarrow$ $Ru$} &	37.44 {\footnotesize ($\pm$ 1.76)} * & \underline{50.76} {\footnotesize ($\pm$ 10.19)} &	\textbf{52.41} {\footnotesize ($\pm$ 8.66)} \\
\hline
\textbf{$En$ $\rightarrow$ $Es$} &	\underline{54.42} {\footnotesize ($\pm$ 2.44)}* & 42.79 {\footnotesize ($\pm$ 4.43)} &	\textbf{58.41} {\footnotesize ($\pm$ 2.84)} \textsuperscript{\textdagger} \\
\textbf{$Ru$ $\rightarrow$ $Es$} &	28.20 {\footnotesize ($\pm$ 4.87)} * & \underline{55.03} {\footnotesize ($\pm$ 5.68)} &	\textbf{63.13} {\footnotesize ($\pm$ 1.89)} \textsuperscript{\textdagger} \\
\hline
\end{tabular}
\caption{Cross-lingual F1 scores with Rest16.}
\label{tab:cross-lingual}
\end{subtable}

\vspace{0.3cm}

\centering
\begin{subtable}[!htb]{0.8\textwidth}
\centering
\small
\begin{tabular}{l|c|c|c}
\hline
\multirow{1}{*}{\textbf{Train $\rightarrow$ Test}} & \multicolumn{1}{c|}{\textbf{SPAN-MBERT}} & \multicolumn{1}{c|}{\textbf{mBART Non-Masked}} & \multicolumn{1}{c}{\textbf{mBART Masked}} \\ 
\hline
\textbf{Rest16$_{En}$\hspace{1.5mm}$\rightarrow$ Lap14$_{En}$} &	31.32 {\footnotesize ($\pm$ 1.74)}* & \underline{41.04} {\footnotesize ($\pm$ 3.61)} &	\textbf{41.58} {\footnotesize ($\pm$ 3.62)}\\
\textbf{MAMS$_{En}$         $\rightarrow$ Lap14$_{En}$}  &	\underline{31.57} {\footnotesize ($\pm$ 2.71)} &  31.23 {\footnotesize ($\pm$ 3.03)}	& \textbf{31.97} {\footnotesize ($\pm$ 2.95)}\\
\hline
\textbf{Lap14$_{En}$\hspace{2mm}$\rightarrow$ Rest16$_{En}$}  &	42.06 {\footnotesize ($\pm$ 2.71)}* & \underline{55.28} {\footnotesize ($\pm$ 5.01)} &	\textbf{57.49} {\footnotesize ($\pm$ 3.07)}\\
\textbf{MAMS$_{En}$         $\rightarrow$ Rest16$_{En}$}      &	\textbf{56.04} {\footnotesize ($\pm$ 1.30)}* &  \underline{50.06} {\footnotesize ($\pm$ 2.12)} &	49.52 {\footnotesize ($\pm$ 1.98)} \\
\hline
\textbf{Rest16$_{En}$\hspace{1.5mm}$\rightarrow$ MAMS$_{En}$}  &	32.32 {\footnotesize ($\pm$ 2.00)}* &  \textbf{36.10} {\footnotesize ($\pm$ 0.93)} &	\underline{35.96} {\footnotesize ($\pm$ 0.89)} \\
\textbf{Lap14$_{En}$\hspace{2mm}$\rightarrow$ MAMS$_{En}$}  &	23.57 {\footnotesize ($\pm$ 2.19)} &  \underline{29.07} {\footnotesize ($\pm$ 3.30)} &	\textbf{30.26} {\footnotesize ($\pm$ 2.13)} \\
\hline
\end{tabular}
\caption{Cross-domain F1 scores.}
\label{tab:cross-domain}
\end{subtable}

\vspace{0.3cm}
\centering
\begin{subtable}[!htb]{0.8\textwidth}
\centering
\small
\begin{tabular}{l|c|c|c}
\hline
\multirow{1}{*}{\textbf{Train $\rightarrow$ Test}} & \multicolumn{1}{c|}{\textbf{SPAN-MBERT}} &  \multicolumn{1}{c|}{\textbf{mBART Non-Masked}} & \multicolumn{1}{c}{\textbf{mBART Masked}} \\ 
\hline
\textbf{Rest16$_{Es}$   $\rightarrow$ Lap14$_{En}$} &	28.52 {\footnotesize ($\pm$ 2.72)}* & \underline{33.84} {\footnotesize ($\pm$ 10.89)} &	\textbf{35.91} {\footnotesize ($\pm$ 9.73)} \\
\textbf{Rest16$_{Ru}$   $\rightarrow$ Lap14$_{En}$}  &	16.80 {\footnotesize ($\pm$ 1.84)}* & \underline{39.56} {\footnotesize ($\pm$ 2.31)} &	\textbf{40.32} {\footnotesize ($\pm$ 2.71)} \\
\hline
\textbf{Lap14$_{En}$    $\rightarrow$ Rest16$_{Es}$} &	\underline{42.26} {\footnotesize ($\pm$ 4.22)} & 36.06 {\footnotesize ($\pm$ 9.09)} &	\textbf{47.42} {\footnotesize ($\pm$ 5.14)} \\
\textbf{MAMS$_{En}$    $\rightarrow$ Rest16$_{Es}$} &	\textbf{47.33} {\footnotesize ($\pm$ 1.26)}* & 20.01 {\footnotesize ($\pm$ 2.54)} &	\underline{39.20} {\footnotesize ($\pm$ 3.32)} \textsuperscript{\textdagger} \\
\hline
\textbf{Lap14$_{En}$    $\rightarrow$ Rest16$_{Ru}$}  &	31.58 {\footnotesize ($\pm$ 7.76)}* & \underline{42.27} {\footnotesize ($\pm$ 8.02)}	& \textbf{42.57} {\footnotesize ($\pm$ 11.01)} \\
\textbf{MAMS$_{En}$    $\rightarrow$ Rest16$_{Ru}$} &	30.75 {\footnotesize ($\pm$ 3.57)} & \underline{34.16} {\footnotesize ($\pm$ 3.72)} &	\textbf{39.42} {\footnotesize ($\pm$ 2.41)} \\
\hline
\textbf{Rest16$_{Es}$    $\rightarrow$ MAMS$_{En}$}  &	\underline{28.78} {\footnotesize ($\pm$ 1.39)} & 25.37 {\footnotesize ($\pm$ 7.25)} &	\textbf{29.05} {\footnotesize ($\pm$ 6.77)} \textsuperscript{\textdagger}  \\
\textbf{Rest16$_{Ru}$    $\rightarrow$ MAMS$_{En}$}	& 14.81 {\footnotesize ($\pm$ 3.23)}* &  \underline{31.50} {\footnotesize ($\pm$ 1.51)} &	\textbf{32.35} {\footnotesize ($\pm$ 1.33)}	\\
\hline
\end{tabular}
\caption{Cross-domain and cross-lingual F1 scores.}
\label{tab:cross-domain-lingual}
\end{subtable}

\caption{Results of experiments performed under various settings. * SPAN-MBERT statistically significantly different from mBART Non-Masked. \textsuperscript{\textdagger} mBART Masked statistically significantly different from mBART Non-Masked. For every setting, the highest model score is bolded and the $2^{nd}$ highest model score is underlined. }
\end{table*}

\subsection{Cross-Lingual}
Table \ref{tab:cross-lingual} presents the cross-lingual results. Except in a case involving testing in Spanish, mBART Non-Masked has better performance than SPAN-MBERT. When testing in Spanish, the mBART Masked model provides a significant improvement (up to 16\%) over the Non-Masked model.

\subsection{Cross-Domain}
Table \ref{tab:cross-domain} presents the cross-domain results. The mBART Non-Masked model performs better than SPAN-MBERT, except in cases involving MAMS$_{En}$ in either test or train. In those cases, three times out of four, SPAN-MBERT does equal or better than the mBART Non-Masked model. In none of the settings is the mBART Masked model statistically significantly better than the mBART Non-Masked model. 

\subsection{Cross-Lingual and Cross-Domain}

This is an extreme setting which combines the cross-lingual and cross-domain settings. Table \ref{tab:cross-domain-lingual} shows the results. In 50\% of the cases, the mBART Non-Masked model does better than the SPAN-MBERT model. In the rest, the SPAN-MBERT performs better or equal to the mBART Non-Masked model. Only in the cases involving both MAMS$_{En}$ and Rest16$_{Es}$ for training/testing or testing/training, does the mBART Masked model have a statistically significant improvement over the Non-Masked model (up to 20\%).

\section{Discussion}

In our experiments, we find that in most cases, the mBART Non-Masked generative model performs better or equal to the SPAN-MBERT discriminative model. But, in 36\% cases, SPAN-MBERT does better or equal to the mBART Non-Masked model.

In several cases involving MAMS$_{En}$, SPAN-MBERT performs better than the mBART Non-Masked model. This includes the monolingual in-domain setting for MAMS$_{En}$, on which generative models were not evaluated, in prior work. MAMS$_{En}$ is a tougher task, and a generative model might need more data to perform well. This can be attributed to the fact that generative models have a challenging task: learning a joint probability over all words. This is in contrast to discriminative models which need only learn a small number of decision boundaries. This intuition is supported by existing literature \cite{ng2002discriminative}.

Based on the results and error analysis (Appendix \ref{cross:ea}), we find that the mBART Non-Masked generative model is unable to understand that it needs to extract words from the input sentence. This leads to errors in cross-lingual and cross-lingual/domain cases. When trained with a language X, and tested with a language Y, the model generates the aspect in language X, instead of language Y.

This problem for cross-lingual and cross-lingual/domain settings especially exists for cases involving English for training/testing and Spanish for testing/training. However, the problem is less pronounced with cases involving Russian. Since the transformer model breaks words into sub-words before tokenization, English and Spanish words might be treated in the same way due to their common sub-words. Russian, on the other hand, has less common sub-words, making it easier for the model to distinguish between Russian and English words. This problem always exists in cases involving MAMS and Rest16 datasets together (in English and Spanish cross-lingual/domain tests). Since both these datasets also belong to the restaurant domain, it is tougher for the model to distinguish between the training and testing datasets. On the other hand, this is less of a problem when using Laptop datasets along with MAMS or Rest16 (in English and Spanish cross-lingual/domain tests) because the model is able to distinguish between the laptop domain and the restaurant domain. Generating out-of-sentence words is not a problem with discriminative models because they are restricted to only the words from the input sentence.

In cross-lingual and cross-domain/lingual cases, the mBART Masked model provides dramatic performance improvements over the mBART Non-Masked model (up to 20\%). This is attributable to the masked model not facing the out-of-sentence aspect generation problem discussed earlier. mBART Masked model always does either better or equal to the Non-Masked model, with it being better than mBART Non-Masked in 16\% cases.

\section{Conclusions}

In this work, we compared two ABSA model types (discriminative and generative) in terms of performance differences by considering a state-of-the-art model for each model type as a representative. We reasoned about these differences in a manner generalizable to discriminative and generative ABSA model types, and not specific to their representative models. Previous studies showed that generative models achieve higher results than the discriminative ones across almost all English ABSA datasets. However, the results in our study demonstrated that generative models can perform worse than the discriminative ones in many of the proposed scenarios, namely, cross-lingual, cross-domain, and cross-lingual and domain. These results argue against adopting generative models as the defacto standard for all ABSA tasks as discriminative models are more accurate in some settings.

We propose a simple modification to the generative model wherein we constrain the decoding strategy. This constrained methodology often leads to significant improvements (of up to 20\%) in the performance of the generative model in cases involving different train and test languages.

In the future, we plan to study the models in other scenarios like conflicting polarities (aspects with both positive and negative polarities).

\section*{Limitations}

\begin{itemize}
    \item We have evaluated transformer based models which are prone to degenerate runs and give high variance results. We have reduced the effect of degenerate runs by using a trimmed means significance test (Yuen-Welch). Moreover, we have used 10 random seeds to minimise variance, but still the variance in model performances is high. We note that our analysis uses statistical significance testing and we only form conclusions from the results which are found to be statistically significant (despite the observed high variance).
    \item Hallucination in natural language generation is a widely studied problem in NLP, and it can be solved using multiple techniques \cite{ji2022survey}. Here, we have demonstrated the effectiveness of a masking based solution, but other solutions are also possible and should be explored.
    \item We use the mBART LLM for our experiments which requires a significant amount of computational resources for training. This leads to a high cost both financially and environmentally \cite{strubell-etal-2019-energy}.
\end{itemize}

\bibliography{anthology,custom}
\bibliographystyle{acl_natbib}

\newpage

\appendix

\section{Model Details}

In the SPAN-MBERT model, we use the multilingual BERT model from Google\footnote{\url{https://github.com/google-research/bert/blob/master/multilingual.md}}.

In both mBART models, we use the multilingual BART-large model from Huggingface\footnote{\url{https://huggingface.co/facebook/mbart-large-cc25}}.

\section{Text Normalisation} \label{norm_output}

Prior to evaluation, the model outputs and the gold data are normalised. We remove punctuation marks such as ",", ".", "\textquotedblright" from the sentences, lower-case and lemmatise the words, and remove common stop words. This is because the generative model often generates a different variant of a term, e.g. plural or singular. This idea of normalising the generated output is similar to \citet{zhang-etal-2021-towards-generative}, where Levenshtein distance is used to align the generated aspect words with the closest words existing in the original sentence. Compared to this, our normalisation process followed by an exact matching is stricter. Levenshtein distance may align the model's predictions with unrelated words in the original sentence. For example, if a generated word - "salmon", has the least distance with the word "not" out of all the words in the original sentence, then "salmon" can get aligned to "not", as is mentioned by \citet{zhang-etal-2021-towards-generative}, which is a loose matching. 

Once model outputs and gold data are normalised, the predicted aspect-polarity terms and the corresponding gold aspect-polarity terms are compared using an exact match. We consider a hit only if both the aspect term and the polarity term match. We use the standard evaluation metrics for calculating ABSA scores, which are Micro- Precision, Recall and F1. We use the evaluation code released by \citet{li2019unified}\footnote{\url{http://github.com/lixin4ever/E2E-TBSA}}.

\section{Error Analysis} \label{cross:ea}

We conducted an error analysis on the outputs of the models to better understand the cases where they fail.

For the discriminative model (SPAN-MBERT), we found that in a large number of the error cases, the model did not predict any aspect term at all. This means that the SPAN-MBERT model was not able to confidently identify any possible aspect term spans, as it uses thresholds (representing confidence) for prediction scores. For example, in the following sentence the model fails to predict an aspect term: "Not the biggest portions but adequate.". We also found several cases where the model correctly identifies the aspect term but misclassifies the sentiment, such as for the sentence "i am never disappointed with there [sic] food."; it gives "food" a negative sentiment instead of a positive. Here, the underlying language model (mBERT) did not understand the word "never", and instead understood the sentiment from "disappointed" which has negative connotations. It has been shown that language models like BERT misunderstand some negations \cite{kassner-schutze-2020-negated}. A significant number of errors are because the predicted and gold aspect spans only have a partial overlap. This can be seen in cases such as "La atención del personal impecable." ("The attention of the impeccable staff.") where the predicted aspect term is "personal" ("staff") instead of "atención del personal" ("attention of the staff").

As in the discriminative model, in both generative models (mBART Non-Masked and mBART Masked) we saw several cases where the predicted aspect span is only partially correct. For instance, in the sentence "Great draft and bottle selection and the pizza rocks.", the predicted entities include "bottle selection" instead of "draft and bottle selection". We note that such predictions would not have been considered errors if we had used partial matching explained earlier in \ref{norm_output}.

In the mBART Non-Masked model, other notable cases included those where an aspect similar to the true aspect is predicted. For example, for the sentence - "The best calamari in Seattle!", the mBART Non-Masked model generated "salmon" as an aspect term instead of "calamari". This shows that the language model understood the similarity between calamari and salmon, however it did not understand that for the task it was supposed to predict a word from the input sentence itself, and not make such inferences. Similarly, in cross lingual experiments, we found that the model would predict the aspect term in the training language, instead of the test language. For example, when training on Rest16$_{En}$ and testing on Rest16$_{Es}$, the model tends to predict "restaurant" instead of "restaurante", "meal" instead of "comida", "service" instead of "servicio", "place" instead of "sitio", "precio" instead of "price". These are all English translations of the required Spanish aspect terms. This again implies that the model is unable to understand that it is supposed to predict a word from the input sentence itself.

\section{Dataset Details} \label{app:dataset_details}

For SemEval datasets, since the validation sets are not given, we sampled 10\% of the training dataset to use for validation. The datasets we considered vary in terms of the type of content and the training set size. Table \ref{table_data_stats} presents the datasets' statistics after cleaning and sampling.

We followed existing work \cite{tian-etal-2021-aspect, tang-etal-2016-aspect} in removing sentences with no opinions (not useful for the considered ABSA task), as well as sentences having aspect terms with a "conflict" sentiment polarity, from the dataset. This is to prevent a class imbalance problem, as there are very few instances of "conflict", compared to other polarities.

\begin{table}[!tb]
\small
\caption{\label{table_data_stats}Datasets' statistics - Count of aspects with sentiment polarities for the cleaned datasets. Multiple aspects can exist in single record}
\centering
\begin{tabular}{l|l|r|r|r}
\textbf{Datasets} & \textbf{Data Split} & \textbf{\#Pos} & \textbf{\#Neg} & \textbf{\#Neu} \\
\hline
\multirow{3}{*}{\textbf{Rest16$_{en}$}} & Train & 1028 & 380 & 54 \\
 & Val & 130 & 32 & 6 \\
 & Test & 427 & 119 & 28 \\
\hline
\multirow{3}{*}{\textbf{Rest16$_{es}$}} & Train & 972 & 338 & 72 \\
 & Val & 101 & 46 & 5 \\
 & Test & 420 & 142 & 29 \\
\hline
\multirow{3}{*}{\textbf{Rest16$_{ru}$}} & Train & 2044 & 411 & 188 \\
 & Val & 223 & 56 & 23 \\
 & Test & 608 & 193 & 85 \\
\hline
\multirow{3}{*}{\textbf{Lap14}} & Train & 895 & 799 & 414 \\
 & Val & 99 & 71 & 50 \\
 & Test & 341 & 128 & 169 \\
\hline
\multirow{3}{*}{\textbf{MAMS$_{En}$}} & Train & 3382 & 2769 & 5042 \\
 & Val & 403 & 325 & 605 \\
 & Test & 400 & 330 & 607 \\

\hline
\end{tabular}
\end{table}

\section{Training Details}

For fine tuning the transformer based ABSA model, we use 1 NVIDIA V100 GPU, 6 CPU cores with 4 GB memory per core. We run training jobs with a 71 hour time limit.

\end{document}